%% file: egpaper_for_review.tex
\documentclass[10pt,twocolumn,letterpaper]{article}

\usepackage{iccv}
\usepackage{times}
\usepackage{epsfig}
\usepackage{caption}
\usepackage{graphicx}
\usepackage{amsmath}
\usepackage{amssymb}
\usepackage{booktabs}
\usepackage[dvipsnames]{xcolor}
\usepackage{tikz}
\usepackage{comment}
\usepackage{xcolor,colortbl}
\usepackage{color}
\usepackage{booktabs}
\usepackage{pifont}
\usepackage{amsfonts}
\usepackage{bbm}
\usepackage{soul}
\usepackage{float}
\usepackage{multirow}
\usepackage{array,makecell}
\usepackage{arydshln}
\usepackage{textcomp}
\usepackage[sort&compress,numbers,square,comma]{natbib}
\usepackage[accsupp]{axessibility}

\usepackage[pagebackref=true,breaklinks=true,letterpaper=true,colorlinks,bookmarks=false]{hyperref}
\definecolor{Blue}{rgb}{0.88,1,1}
\definecolor{Gray}{gray}{0.85}

\newcommand*{\belowrulesepcolor}[1]{%
  \noalign{%
    \kern-\belowrulesep
    \begingroup
      \color{#1}%
      \hrule height\belowrulesep
    \endgroup
  }%
}
\newcommand*{\aboverulesepcolor}[1]{%
  \noalign{%
    \begingroup
      \color{#1}%
      \hrule height\aboverulesep
    \endgroup
    \kern-\aboverulesep
  }%
}
\usepackage[capitalize]{cleveref}
\crefname{section}{Sec.}{Secs.}
\Crefname{section}{Section}{Sections}
\Crefname{table}{Table}{Tables}
\crefname{table}{Tab.}{Tabs.}
\newcommand{\cmark}{\ding{51}}%
\newcommand{\xmark}{\ding{55}}%
\definecolor{Gray}{gray}{0.9}
\iccvfinalcopy %

\ificcvfinal\fi

\begin{document}

\title{SparseDet: Improving Sparsely Annotated Object Detection with Pseudo-positive Mining}

\newcommand\blfootnote[1]{%
  \begingroup
  \renewcommand\thefootnote{}\footnote{#1}%
  \addtocounter{footnote}{-1}%
  \addtocounter{Hfootnote}{-1}%
  \endgroup
}

\author{
Saksham Suri\thanks{First two authors contributed equally}$\text{ }^{1}$\\
{\tt\small sakshams@cs.umd.edu}
\and
Saketh Rambhatla\footnotemark[1]$\text{ }^{1}$\\
{\tt\small rssaketh@umd.edu}
\and
Rama Chellappa$^{1,2}$\\
{\tt\small rchella4@jhu.edu}
\and
Abhinav Shrivastava$^{1}$\\
{\tt\small abhinav@cs.umd.edu}
\and
{University of Maryland, College Park$^{1}$}\quad\quad
{Johns Hopkins University$^{2}$}
}

\maketitle

\ificcvfinal\thispagestyle{empty}\fi

\begin{abstract}
Training with sparse annotations is known to reduce the performance of object detectors. Previous methods have focused on proxies for missing ground truth annotations in the form of pseudo-labels for unlabeled boxes. We observe that existing methods suffer at higher levels of sparsity in the data due to noisy pseudo-labels. To prevent this, we propose an end-to-end system that learns to separate the proposals into labeled and unlabeled regions using Pseudo-positive mining. While the labeled regions are processed as usual, self-supervised learning is used to process the unlabeled regions thereby preventing the negative effects of noisy pseudo-labels. This novel approach has multiple advantages such as improved robustness to higher sparsity when compared to existing methods. We conduct exhaustive experiments on five splits on the PASCAL-VOC and COCO datasets achieving state-of-the-art performance. We also unify various splits used across literature for this task and present a standardized benchmark. On average, we improve by $2.6$, $3.9$ and	$9.6$ mAP over previous state-of-the-art methods on three splits of increasing sparsity on COCO. 
Our project is publicly available at this \href{https://cs.umd.edu/~sakshams/SparseDet}{website}.

\end{abstract}

\input{introduction}

\section{Related Work}
\label{sec:related_work}

\noindent\textbf{Semi-supervised object detection: }Semi-supervised object detection (SSOD) is an active field that also deals with training object detectors with missing annotations.
Existing works on semi-supervised object detection, have focused mainly on consistency regularization~\cite{jeong2019consistency,tang2021proposal} or pseudo-labeling-based approaches~\cite{sohn2020simple,liu2021unbiased,xu2021end,yang2021interactive}. 
The main idea behind these approaches is to perturb the images, or features, and apply a consistency regularization loss to enforce consistency between the predictions using a student teacher framework. However, typical SSOD methods assume a small exhaustively labeled set and a large unlabeled set for training. This is different from Sparsely annotated object detection (SAOD), which assumes a large training set which is sparsely labeled.

\noindent\textbf{Sparsely annotated object detection: } 
One of the initial works addressing this problem, by Niitani \etal~\cite{niitani2019sampling}, proposes utilizing logical relations between the co-occurrences of objects and pseudo-labeling.
Yoon \etal~\cite{yoon2021semi} use object tracking to densely label objects across sparsely annotated frames along with single stage detectors to mitigate the negative effects of missing annotations. 
Wu \etal~\cite{wu2018soft} propose a re-weighting approach where the gradients corresponding to region of interest are weighed as a function of overlap with ground truth instances.
Improving upon  the previous work, Zhang \etal~\cite{zhang2020solving} propose an automatic re-calibration strategy for single stage detectors where the negative branch is changed to take into account low confidence background predictions which might correspond to missing annotations.

Finally, one of the most recent works, Co-mining~\cite{Wang2021CominingSL} uses a co-training strategy by using two views of an image and predictions from one view along with the  ground truth as supervision for the other view and vice versa. Our method doesn't solely rely on pseudo-labels and leverages a self-supervised loss to prevent propagating negative gradients to the model due to false negatives.

\noindent\textbf{Fully supervised object detection: } 
After the success of AlexNet~\cite{NIPS2012_4824} on the image classification challenge (ILSVRC 2012)~\cite{ILSVRC15}, research on designing deep neural networks for object detection gathered more interest. First among the successful methods are the two-stage \textbf{Region-based convolutional networks} (R-CNN) family of detectors. Two-stage object detectors consists of 1) a region proposal stage which produces a set of candidate object bounding boxes followed by 2) a classification stage which classifies each candidate region as either belonging to a foreground object or ``background". R-CNN processes a large amount of region proposals by cropping the input image and using a CNN backbone to extract features making it extremely slow.
\textbf{Fast R-CNN}~\cite{Girshick:2015} was proposed to overcome this limitation. Fast R-CNN computes one convolution feature map for the whole image. RoI Pooling was introduced to pool the feature for each region of interest into a fixed spatial dimension. RoI pooling shares the computation among all the region proposals speeding up training and inference. Fully connected layers are applied on the fixed RoI pooled feature maps which are then passed to two sister heads, for classification and bounding box regression. The whole network is trained end-to-end with a multi-task loss avoiding the multi-stage training in R-CNN~\cite{girshick2014rich}. 
While Fast R-CNN improved the efficiency of its predecessor, test time computation bottleneck is still an issue because of the region proposal method employed. \textbf{Faster R-CNN}~\cite{Ren2015FasterRT} proposed a region proposal network (RPN), an elegant solution that trains deep networks to predict region proposals for practically no additional computational overhead. RPN shares convolutional layers with Fast R-CNN~\cite{Girshick:2015} and at test time the cost of generating proposals is minimal. 
Faster R-CNN paved the way for more sophisticated and efficient two stage detection architectures~\cite{dai2016object,Lin2017FeaturePN,analysissnip2017,sniper2018,najibi2019autofocus}. Faster R-CNN has also been extended to achieve state-of-the-art instance segmentation~\cite{8237584}, panoptic segmentation~\cite{DBLP:conf/cvpr/KirillovGHD19,DBLP:conf/cvpr/KirillovHGRD19}, and 3D mesh generation~\cite{meshrcnn} \etc. A few limitations of two stage object detectors and anchor boxes has also inspired the family of single-stage~\cite{liu2016ssd, Lin2017FocalLF, redmon2015unified} and anchor-free object detection systems~\cite{tian2019fcos, Duan_2019_ICCV}.

\section{Approach}\label{sec:approach}
\input{approach.tex}

\section{Experimental Evaluation} \label{sec:experimental_evaluation}

\input{experiments.tex}

\input{conclusion.tex}

\input{acknowledgement.tex}

{\small
\bibliographystyle{unsrtnat}
\bibliography{egbib}
}
\cleardoublepage
\input{iccv_supp.tex}

\end{document}

%% file: introduction.tex
\section{Introduction}
\begin{figure}[!tbh]
    \centering
    \includegraphics[width=\linewidth]{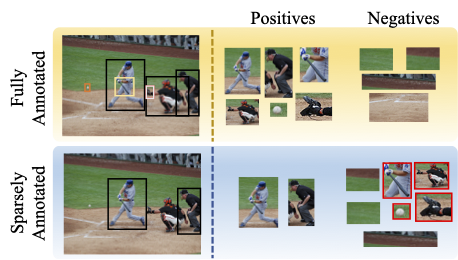}
    \caption{(Top) Most Object Detection datasets have exhaustive annotations for foreground/positives. During training, the unlabeled regions can be safely considered as background/negatives. Sparsely Annotated Object Detection datasets (bottom) have missing annotations. This results in foreground regions (shown in red) being considered as negatives during training, adversely affecting the performance of the classifier.}
    \label{fig:comp_teaser}
\end{figure}
 The performance of object detectors is sensitive to the quality of labeled data~\cite{Chadwick2019TrainingOD,wu2018soft,Li2020TowardsNO}. Existing object detection methods assume that the training data is pristine and a drop in performance is observed if this assumption fails.
 Noise in the data used for training object detectors can arise due to incorrect class labels or incorrect/missing bounding boxes. 
In this work, we deal with the problem of training object detectors with sparse annotations, i.e., missing region or bounding boxes.
This problem is of utmost importance, as obtaining crowd-sourced datasets~\cite{Deng2009ImageNetAL,Lin2014MicrosoftCC} can be expensive and laborious. The alternative is to use computer-assisted protocols to collect annotations which have been shown to be noisy and incomplete~\cite{Kuznetsova2020TheOI}. %
This problem of training object detectors with incomplete bounding box annotations is called Sparsely Annotated Object Detection (SAOD).

To understand why training with sparse annotations is detrimental to the performance, consider the example shown in Figure~\ref{fig:comp_teaser} (top). If the annotation were exhaustive, then the negative samples to the classifier contain \textit{true} background regions. But with sparse annotations, as shown in Figure~\ref{fig:comp_teaser} (bottom), a few positive regions will inevitably be considered as negatives (shown in red), thereby wrongfully penalizing the classifier leading to lower performance. Existing methods\cite{niitani2019sampling,wu2018soft,zhang2020solving,yang2020object,Wang2021CominingSL} prevent this by predicting pseudo-labels and removing the foreground regions from the negatives to the classifier. However, at higher levels of sparsity, the quality of pseudo-labels is greatly affected, resulting in the same problem noted above.

Crowd sourced object detection datasets~\cite{pascal-voc-2007,pascal-voc-2012,Lin2014MicrosoftCC} are ensured to be almost exhaustively labeled. Hence, for SAOD, researchers artificially create sparsely annotated datasets from the original ones. There is no general consensus on the correct way to create the sparsely annotated datasets, \textit{a.k.a.} splits, and hence each method reports results on one or two different splits. Split can be created by considering the dataset as a whole or per image (\ie annotations can be removed by considering all the images or only a single image at once). They can also be created by removing annotations in a class-agnostic or class-aware fashion (\ie remove $p\%$, of annotations per category or across all the categories). These variations result in splits with different data distributions making some harder than the others (refer to Table~\ref{tab:saod_coco}). A proper benchmark that analyzes the performance of SAOD methods across these different types of splits is missing. This makes it difficult to compare methods and assess their effectiveness for a specific use case.

To tackle the issues discussed above, as our first major contribution, we present SparseDet, a novel SAOD framework that achieves state-of-the-art performance across multiple SAOD benchmarks in practice.
SparseDet operates on an image and its augmented counterpart. The combination of features extracted from the two views is used to generate region proposals. Standard detection training methods, consider a region proposal as positive if its intersection over union (IoU) with any ground truth is greater than 0.5 and the rest are treated as negatives. This strategy works when the annotations are exhaustive, which implies that the remaining regions are from background. But due to missing annotations, some of these regions could belong to foreground instances. To prevent considering all region proposals without annotations as negatives, SparseDet partitions all the region proposals into labeled, \textit{unlabeled} and background. The labeled and background regions are processed as usual. Features extracted from unlabeled regions are then trained with a self-supervised loss. Previous approaches like Co-Mining~\cite{Wang2021CominingSL}, consider two partitions, labeled and background, and generate pseudo-labels. This is a disadvantage at high sparsity as the generated pseudo-labels can be very noisy. The self-supervised loss in our approach enforces consistency between the features of the two views for the unlabeled regions and prevents penalizing the classifier due to false negatives.

Our second major contribution is unifying evaluation. The standard practice is to simulate sparsely annotated training data on COCO~\cite{Lin2014MicrosoftCC} and  PASCAL-VOC~\cite{pascal-voc-2007} train sets and evaluate them on their corresponding standard validation set. As discussed above, a survey of recent SAOD approaches~\cite{niitani2019sampling,wu2018soft,zhang2020solving,yang2020object,Wang2021CominingSL,Yan2020UniversalLD,Li2020ANL,Zhang2020BeyondWS,Yan2021LearningFM} reveals that there are at least \textbf{\textit{five}} different ways to create splits, each differing in the strategy (refer to Section~\ref{subsec:split}) used to achieve the desired level of sparsity. However, these splits have not been made public, making it difficult to replicate results for comparison. Additionally, each of these strategies has a different property for simulating sparse data, \eg, different distribution of annotations per class, resulting in a different training set.
As a result, methods trained on different splits cannot be compared with one another. To mitigate these issues we standardize the generation of these splits that enables the evaluation of any SAOD methods on all of them for fair comparison.
Additionally, we propose a new benchmark that assesses the semi-supervised learning capabilities of SAOD methods, \ie, leveraging unlabeled data to improve performance. We present our approach as a baseline. We will make the data for the new benchmark along with all the SAOD splits public to facilitate future research. 
We briefly summarize our contributions below.

\begin{itemize}
    \setlength\itemsep{0.1em}
    \item We propose a novel formulation, SparseDet, for SAOD which is an end-to-end approach that identifies labeled, unlabeled and background regions and deals with them in appropriate manner.
    \item We show state-of-the-art performance on sparsely annotated object detection across various splits. On average, we improve by $2.6$, $3.9$ and $9.6$ mAP over previous state-of-the-art methods on three splits of increasing sparsity on COCO. 
    \item We standardize the experimental setup for SAOD by evaluating methods on all the splits to facilitate future research. Additionally, we propose a new benchmark that evaluates the semi-supervised learning capabilities of SAOD methods.
\end{itemize}

In Section~\ref{sec:related_work}, we discuss the related works on SAOD and related fields. We describe our approach in detail in Section~\ref{sec:approach}. We describe our experimental setup and present results in Section~\ref{sec:experimental_evaluation}, and conclude in Section~\ref{sec:conclusion}.

%% file: approach.tex
\begin{figure*}[!t]
\centering
\includegraphics[width=\linewidth]{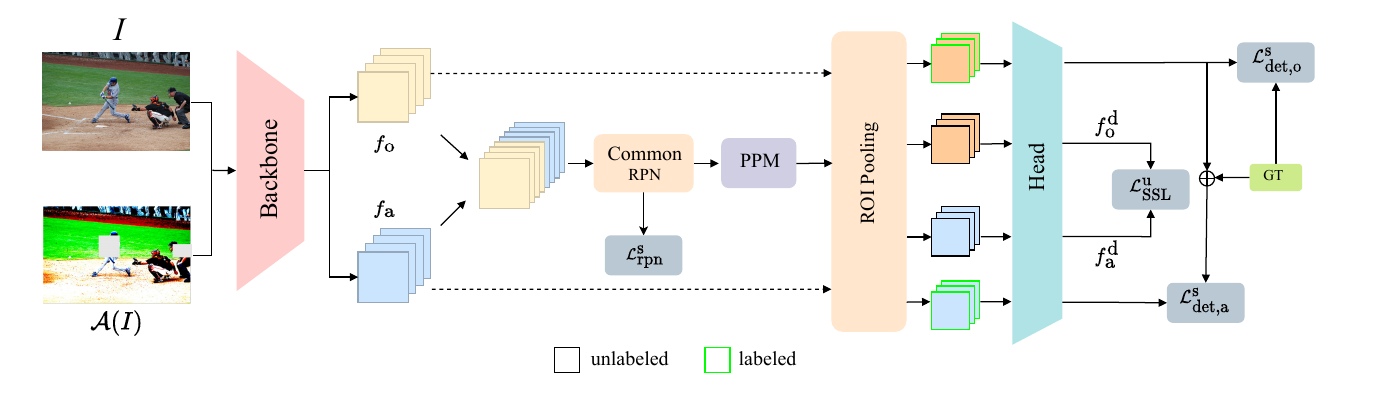}
  \caption{Illustration of SparseDet for sparsely annotated object detection. Following feature extraction from the original and augmented image, a common set of proposals is generated by the common RPN (C-RPN). Using an end to end approach, we identify and mine proposals corresponding to missing annotations using pseudo positive mining (PPM). We train the network end-to-end using a combination of supervised and self-supervised losses. The unlabeled instances (black) are supervised with a self-supervised loss and the labeled instances (green) are supervised with ground truth annotations.}
\label{fig:pipeline}\vspace{-1.5em}
\end{figure*}

We present SparseDet for Sparsely Annotated Object Detection. Given $N$ images, we denote the set of labeled regions in the dataset as $\mathcal{B}_\text{l} = \{b_i, c_i\}_{i=1}^{\text{N}_\text{l}}$ where $(b_*, c_*)$ are the bounding box and class labels respectively. 
The unlabeled regions are denoted as $\mathcal{B}_{\text{ul}} = \{b_k\}_{k=1}^{\text{N}_{\text{ul}}}$ and the background regions as $\mathcal{B}_{\text{bg}} = \{b_q\}_{q=1}^{\text{N}_{\text{bg}}}$.
Note that unlike existing SAOD approaches, we do not assume images to contain at least one labeled region in this work.  The unlabeled $\mathcal{B}_{\text{ul}}$ and background $\mathcal{B}_{\text{bg}}$ sets are not known a-priori.
\subsection{Overview}
The proposed approach is shown in Figure~\ref{fig:pipeline} and consists of a backbone network that extracts features from the original and augmented views of an image. The common RPN (C-RPN), concatenates the features, to generate a set of region proposals. A region proposal $b$  can belong to one of three groups, namely 1) labeled regions $b\in\mathcal{B}_\text{l}$, 2) unlabeled foreground regions $b\in\mathcal{B}_{\text{ul}}$, or 3) background regions $b\in\mathcal{B}_{\text{bg}}$. For a given set of ground-truth annotations, the first group, i.e. labeled regions can be automatically identified. The problem then reduces to identifying and separating the second group i.e. the unlabeled regions, from the background regions. 
Given all the region proposals, a pseudo-positive mining (PPM) step identifies the unlabeled regions and segregates them from the background regions. The labeled and unlabeled regions are trained using supervised and self-supervised losses respectively. 
We describe each stage in detail below.
\subsection{Feature Extraction} %
Given an image $I$, an augmented version of $I$ denoted as $\mathcal{A}(I)$ is computed. In this work, we use random contrast, brightness, saturation, lighting and bounding box erase in a cascaded fashion to generate $\mathcal{A}(I)$. A backbone network is employed to extract two features, $f_\text{o}$ and $f_\text{a}$, from $I$ and $\mathcal{A}(I)$ respectively.
\subsection{Common RPN (C-RPN)}
Two stage object detectors~\cite{Lin2017FeaturePN,Ren2015FasterRT,He2020MaskR} use a region proposal network (RPN)~\cite{Ren2015FasterRT} to generate regions of interest (RoIs). We propose C-RPN which concatenates $f_\text{o}$ and $f_\text{a}$ to obtain the RoIs. This is different from existing approaches that use $f_\text{o}$ and $f_\text{a}$ to generate two separate sets of RoIs. Operating on two sets increases the difficulty of identifying the labeled, unlabeled and background regions which are required for subsequent stages. 
The sparse ground truth is used as supervision to train the C-RPN with a binary cross entropy loss for foreground classification and smooth L1~\cite{fastrcnn} loss for bounding box regression as shown below:
\begin{gather}
    \mathcal{L}_{\text{rpn}}^{\text{s}}\left(\mathbf{c},\mathbf{b}, \mathbf{c^*},\mathbf{b^*}\right) = \mathcal{L}_{\text{bce}}\left(\mathbf{c},\mathbf{c^*}\right) + \lambda\mathcal{L}_{\text{reg}}\left(\mathbf{b}, \mathbf{b^*}\right)
\end{gather}
where $\mathbf{c}$, $\mathbf{b}$ are the objectness scores and bounding boxes, and $\mathbf{c^*}$, $\mathbf{b^*}$ are the corresponding ground truth.

\subsection{Pseudo Positive Mining (PPM)}\label{sec:ppm}
Given the RoIs from C-RPN, the next step is to seggregate the unlabeled regions from the labeled and background regions. 
A standard practice in training detectors is to consider all proposals with an objectness score greater than $\tau_{\text{obj}}$ and IoU greater $\tau_{\text{fg}}$  with any ground truth as foreground and proposals with IoU less than $\tau_{\text{bg}}$ as background. In the presence of missing annotations, detectors are wrongly penalized because the IoU of RoIs corresponding to missing annotations with any ground truth is less than $\tau_{\text{bg}}$. To avoid this, we mine “potential” foregrounds and consider them as unlabeled regions ($\mathcal{B}_{\text{u}}$). The proposed PPM module is based on our observation that when trained with sparse annotations, the RPN can reliably distinguish foreground from background regions. We pick all RoIs that have an objectness score greater than $\tau_{\text{ppm}}$ and IoU less than $\tau_{\text{bg}}$ with any ground truth as the unlabeled regions.  The remaining RoIs are assigned as background.
\begin{table*}[t]
 \setlength{\cmidrulewidth}{0.01em}
\renewcommand{\tabcolsep}{12pt}
\renewcommand{\arraystretch}{0.8}
\newcolumntype{a}{>{\columncolor{Blue}}c}
\centering
\setlength{\extrarowheight}{0.75ex}
\caption{\textbf{Sparsely annotated object detection} results on three splits of COCO dataset. ``Oracle" corresponds to training models using all annotations. Results are reported on the COCO validation set using AP[0.50:0.95]. }\label{tab:saod_coco} %
\resizebox{\linewidth}{!}{
\begin{tabular}{lccaccaccac}
\toprule
 \textbf{Method} &  \multicolumn{3}{c}{\textbf{Split-1}}&  \multicolumn{3}{c}{\textbf{Split-2}} & \multicolumn{3}{c}{\textbf{Split-3}} & \multirow{2}{*}{100\%} \\
 \cmidrule[\cmidrulewidth](l){2-4}
 \cmidrule[\cmidrulewidth](l){5-7}
 \cmidrule[\cmidrulewidth](l){8-10}

 &  30\% & 50\% & 70\% & 30\% & 50\% &70\% & 30\% & 50\% & 70\% \\
 \midrule

\belowrulesepcolor{Gray}
\rowcolor{Gray}Oracle &  & & & & & & & & & 40.91\\
\aboverulesepcolor{Gray}
 \midrule
 Pseudo Label~\cite{niitani2019sampling} & - & 27.50 & - & - & - & - & - & - & - & -\\
 BRL~\cite{zhang2020solving} & - & 32.70 & - & - & - & - & - & - & - & -\\
 Co-mining~\cite{Wang2021CominingSL}  & 36.35 & 32.84 & 24.93 & 36.72 & 33.04 & 24.83 & 36.76 & 32.54 & 24.96 & - \\
 
 Ours & \textbf{38.22}&	\textbf{35.92}&	\textbf{32.68}&	\textbf{39.76}&	\textbf{36.94}&	\textbf{35.33}&	\textbf{39.56}&\textbf{37.15}&	\textbf{35.48}& -\\
 
\bottomrule
\end{tabular}
  }
  \vspace{-1em}

\end{table*}
\subsection{Losses} %
The pseudo positive mining step segregates the RoIs into labeled, unlabeled and background regions. 
An RoI pooling layer~\cite{Ren2015FasterRT} extracts RoI pooled features for the labeled and background regions using the feature $f_\text{o}$.  The detection head processes the RoI pooled features to predict class-wise probabilities and bounding box adjustments for each region. The ground truth is used to supervise the predictions by applying the cross entropy loss for classification and smooth L1~\cite{fastrcnn} for bounding box regression as shown below:
\begin{gather}
    \mathcal{L}_{\text{det},\text{o}}^\text{s}\left(\mathbf{c},\mathbf{b}, \mathbf{c^*},\mathbf{b^*}\right) = \mathcal{L}_{\text{ce}}\left(\mathbf{c},\mathbf{c^*}\right) + \lambda\mathcal{L}_{\text{reg}}\left(\mathbf{b}, \mathbf{b^*}\right)
\end{gather}
where $\mathbf{c}$, $\mathbf{b}$ are the class and bounding box predictions, and $\mathbf{c^*}$, $\mathbf{b^*}$ are the corresponding ground truth class labels and bounding boxes. 

All detections with a score greater than $\tau_{\text{m}}$ are combined with the ground truth followed by a class specific NMS to obtain the merged ground truth. It is ensured that no ground truth annotation is discarded during this step. The labeled and background regions along with $f_\text{a}$ are used to extract RoI pooled features which are then fed to the detector head. The predictions of the detector head are supervised with the merged ground truth with the following losses:
\begin{gather}
    \mathcal{L}_{\text{det},\text{a}}^\text{s}\left(\mathbf{c},\mathbf{b}, \mathbf{c^*},\mathbf{b^*}\right) = \mathcal{L}_{\text{bce}}\left(\mathbf{c},\mathbf{c_m^*}\right) + \lambda\mathcal{L}_{\text{reg}}\left(\mathbf{b}, \mathbf{b_m^*}\right)
\end{gather}
where $\mathbf{c}$, $\mathbf{b}$ are the class and bounding box predictions from the detector head, and, $\mathbf{c_m^*}$, $\mathbf{b_m^*}$ are the corresponding merged ground truths. 

Finally, a class agnostic NMS is performed on the unlabeled regions $\mathcal{B}_{\text{u}}$ (obtained after PPM described in Sec.~\ref{sec:ppm}). The unlabeled regions along with $f_\text{o}$ and $f_\text{a}$ are passed through the ROI pooling layer and the detection head to obtain $f^\text{d}_\text{o}$ and $f^\text{d}_\text{a}$ respectively. A self-supervised loss is applied that enforces the detection head features of the original and augmented regions to be consistent with each other as shown below:
\begin{gather}
    \mathcal{L}_{\text{SSL}}^{\text{u}}\left(f^\text{d}_{\text{a}}, f^\text{d}_{\text{o}}\right) = \|f^\text{d}_{\text{a}} - f^\text{d}_{\text{o}}\|^2_2\,
\end{gather}

The network is trained end-to-end with both supervised and unsupervised losses as shown below:
\begin{gather}
\mathcal{L} = 0.5*\left(\mathcal{L}_{\text{det},\text{o}}^\text{s} + \mathcal{L}_{\text{det},\text{a}}^\text{s}\right) + \mathcal{L}_{\text{rpn}}^\text{s} + \mathcal{L}_{\text{SSL}}^\text{u}
\end{gather}
\noindent\textbf{Discussion: }
In the absence of ground truth annotations, i.e. for completely unlabeled images, the supervised losses ( $\mathcal{L}_{\text{det},\text{o}}^\text{s}$, $\mathcal{L}_{\text{det},\text{a}}^\text{s}$, $\mathcal{L}_{\text{rpn}}^\text{s}$) cannot be computed. Our proposed approach leverages self-supervised consistency loss that does not need ground truth. This helps our approach leverage these unlabeled regions unlike contemporary SAOD methods. We claim that this is the first method to use self-supervised losses for SAOD. Even though Co-Mining~\cite{Wang2021CominingSL} claims to use self-supervised learning it is technically co-learning. Ours is the first approach to use pseudo-labeling and a self-supervised loss to handle the sparse annotations.

%% file: experiments.tex
In this section, we describe the experiments to evaluate our proposed approach. In Sections~\ref{sec:data} and~\ref{subsec:split}, we describe data, splits, and metrics. In Section~\ref{sec:impl}, we mention the implementation details followed by the baselines in Section~\ref{sec:baseline}. We compare with contemporary methods in Section~\ref{sec:sota}, followed by an ablation study in Section~\ref{sec:ablation}.

\begin{table*}[!t]
\begin{minipage}[b]{0.6\linewidth}
\renewcommand{\arraystretch}{1.0}
\newcolumntype{a}{>{\columncolor{Blue}}c}
\centering
\caption{\textbf{Sparsely annotated object detection} results on two splits of PASCAL-VOC dataset. ``Oracle" corresponds to training models using all annotations. Results are reported on VOC 07 test set usiing AP$_{50}$.}\label{tab:saod_voc} %
\resizebox{\linewidth}{!}{
\begin{tabular}{lccacca}
\toprule
 \textbf{Method} & \multicolumn{3}{c}{\textbf{Split-4}}& \multicolumn{3}{c}{\textbf{Split-5}}  \\
 \cmidrule[\cmidrulewidth](l){2-4}
 \cmidrule[\cmidrulewidth](l){5-7}

 &  Easy & Hard & Extreme & 30\% &40\% & 50\% \\
 \midrule

\belowrulesepcolor{Gray}
\rowcolor{Gray}Oracle  & \multicolumn{3}{c}{83.09}&\multicolumn{3}{c}{ 77.47} \\
\aboverulesepcolor{Gray}

 \midrule
 
 BRL~\cite{zhang2020solving}  & 73.50 & 71.70 & 66.20 & - & - & - \\ %
 Co-mining~\cite{Wang2021CominingSL}  & 79.59 & 78.38 & 69.60 & 74.42 & 73.30 & 69.89 \\

 Ours & \textbf{82.15} & \textbf{81.50} & \textbf{75.59}& \textbf{76.84} & \textbf{75.88} & \textbf{74.35}\\
\bottomrule
\end{tabular}
}
\end{minipage}\hfill
\begin{minipage}[b]{0.38\linewidth}
\setlength{\cmidrulewidth}{0.01em}
\renewcommand{\tabcolsep}{8pt}
\renewcommand{\arraystretch}{1.6}
\caption{Results on the proposed \textbf{SSL+SAOD} setup. VOC12~\cite{pascal-voc-2012} is used as the unlabeled data and $p$ is the removal percentage}\label{tab:tssl}
\resizebox{\linewidth}{!}{
\begin{tabular}{@{}lcccccccccccccc@{}}
\toprule
 \textbf{Method} &\multicolumn{3}{c}{\textbf{AP}}  \\
  \cmidrule[\cmidrulewidth](l){2-4}
 & 30\% & 40\% & 50\%\\

 \midrule

 Co-mining~\cite{Wang2021CominingSL} & 46.53 & 45.98 & 43.21\\

 Ours  &   \textbf{48.34} & \textbf{47.82} & \textbf{46.76}\\ %
\bottomrule
\end{tabular}
}
\end{minipage}

\vspace{-0.5em}
\end{table*}

\begin{figure*}[t]
\includegraphics[width=\linewidth]{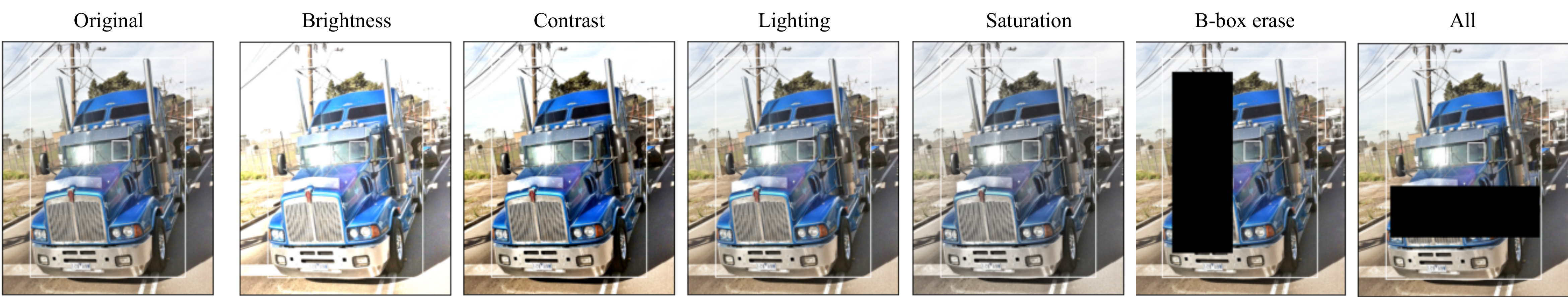}
  \caption{Illustration of the effects of various augmentations used in this work.}
\label{fig:aug}
\vspace{-0.9em}
\end{figure*}

\subsection{Data and Metrics}\label{sec:data}
We conduct all our experiments on the COCO~\cite{Lin2014MicrosoftCC} and PASCAL-VOC~\cite{pascal-voc-2007,pascal-voc-2012} (2007+2012) datasets. The COCO~\cite{Lin2014MicrosoftCC} dataset consists of $118000$ and $5000$ images for training and validation respectively. Experiments on the PASCAL-VOC07~\cite{pascal-voc-2007} are conducted on $5011$ trainval images and performance is computed on $4952$ images of the test set. The PASCAL-VOC12 version consists of $11530$ (trainval) images for training and evaluation is done on the PASCAL-VOC07 test set. Following past literature~\cite{niitani2019sampling,wu2018soft,zhang2020solving,yang2020object,Wang2021CominingSL}, we create five different splits (Section~\ref{subsec:split}) and report results on them. For splits on the COCO~\cite{Lin2014MicrosoftCC} dataset, we use the standard COCO style Average Precision (AP). For splits on the PASCAL-VOC~\cite{pascal-voc-2007} dataset, we use the standard PASCAL-VOC style AP$_{50}$, which is Average Precision computed at an IoU threshold of $0.5$. 

\subsection{SAOD Splits}\label{subsec:split}
An extensive review of sparsely annotated object detection methods reveals that there are atleast five popular types of splits in use for creating training data. Most methods report results on a subset of these making it harder to compare across methods.

We standardize the evaluation of SAOD methods by evaluating exhaustively on all the splits facilitating future research. We briefly describe the splits below. 

\noindent\textbf{Split-1{\normalfont ~\cite{yang2020object,Wang2021CominingSL}}: }In this split, for each object category, $p\%$ of annotations are randomly removed from the COCO~\cite{Lin2014MicrosoftCC} train set and results are reported on the validation set. This split simulates sparsity at dataset level in a class aware fashion. Note, this split can contains images with no annotations. We experiment with $p=\{30, 50, 70\}$.
 
\noindent\textbf{Split-2: }
For each image in the COCO~\cite{Lin2014MicrosoftCC} train set, all annotations of $p\%$ of all categories in that image removed and results are reported on the COCO validation set. This split can be considered as image level and class aware. We experiment with $p=\{30, 50, 70\}$. Note, this split might contain images with no annotations.\\
\noindent\textbf{Split-3: }This split, which uses the COCO 2017~\cite{Lin2014MicrosoftCC} train set for training and the validation set for evaluation, deletes $p\%$ annotations in a class agnostic fashion for each image ensuring at least one annotation. This split is image level but class agnostic. For the experiments, we use $p=\{30, 50, 70\}$. 

\noindent\textbf{Split-4{\normalfont~\cite{zhang2020solving,Wang2021CominingSL}}: }This split requires evaluating models on three different settings namely \textit{easy}, \textit{hard} and \textit{extreme} ensuring at least one annotation per image constructed using PASCAL-VOC 2007+12~\cite{pascal-voc-2007,pascal-voc-2012} trainval set. Results are reported on the PASCAL-VOC 2007~\cite{pascal-voc-2007} test set. For each image in the training set, the easy setting randomly removes one annotation while the \textit{hard} setting randomly removes half of the annotations. The \textit{extreme} setting retains only one annotation per image. All the sets ensure each image consists of atleast one annotation. This split is a small scale version for an image level class agnostic split.

\noindent\textbf{Split-5{\normalfont~\cite{wu2018soft}}: } This split uses the PASCAL-VOC 2007~\cite{pascal-voc-2007} train set for training and the PASCAL-VOC 2007 test set for evaluation and drops $p\%$ annotations per class. For each image in this construction, instances of randomly selected categories are exhaustively annotated while the remaining categories do not have any annotations. This is a small scale version of Split-1 as annotations are dropped in a class aware manner across the full dataset maintaining atleast 1 annotation per image. In this case we use $p=\{30, 40, 50\}$.

\subsection{Implementation details}\label{sec:impl}
For all our experiments, we use a Faster RCNN~\cite{Ren2015FasterRT} architecture with a ResNet-101~\cite{resnet} FPN backbone~\cite{Lin2017FeaturePN} implemented using Detectron2~\cite{wu2019detectron2} framework. For the image augmentations, we apply a For augmentation, a cascade of random contrast, brightness, saturation, lighting and bounding box erase augmentations. The effect of this augmentation is shown in Figure~\ref{fig:aug}. We train all our models with a batch size of $8$ for $270000$ and $18000$ iterations on the COCO and PASCAL-VOC splits respectively with a learning rate of $0.01$. The learning rate is decreased ten fold twice at $\{210000, 250000\}$ and $\{12000, 15000\}$ for COCO and PASCAL-VOC respectively. We adopt a warm-up strategy for $1000$ and $100$ iterations for the COCO and PASCAL-VOC respectively. Following existing detector implementations we set $\tau_{\text{bg}}, \tau_{\text{fg}}$, $\tau_{\text{obj}}$ and $\tau_{\text{ppm}}$ to $0.2, 0.4$, $0.5$ and $0.8$ respectively. We set $\tau_{m}$ to $0.9$. During inference, we compute the backbone features ($f_\text{o}$ and $f_\text{a}$) for both the original and augmented versions of the input to obtain the RoIs and only $f_\text{o}$ is used for the final detections.

\subsection{Baselines}\label{sec:baseline}
We compare our method against Co-Mining~\cite{Wang2021CominingSL}, BRL~\cite{zhang2020solving}, and Pseudo Label~\cite{niitani2019sampling}. 

We choose these methods as they outperform previous approaches for the SAOD task. We use the public implementation of these methods and usa a ResNet 101 FPN~\cite{Lin2017FeaturePN} backbone for all the experiments in order to perform a fair comparison.
 \begin{table*}[!ht]
\begin{minipage}[]{0.52\linewidth}
 \setlength{\cmidrulewidth}{0.01em}
\renewcommand{\tabcolsep}{4pt}
\renewcommand{\arraystretch}{1.5}
\centering
\caption{Ablation of various components of proposed approach.}
\label{tab:cr}
\begin{tabular}{@{}cccccc@{}}
\toprule
 \textbf{C-RPN} & \textbf{PPM}&$\mathbf{\mathcal{L}^{u}_{SSL}}$&$\mathbf{\mathcal{L}^{s}_{det,a}}$&\textbf{AP}& $\Delta$   \\
 \midrule
 \xmark & \xmark & \xmark & \xmark & 41.77& $-$\\
 \xmark & \xmark & \xmark & \cmark & 42.67& 0.90\\
 \cmark & \xmark & \xmark & \cmark & 45.30 & 3.53\\
  \cmark & \cmark & \xmark &\cmark &45.44&	3.67\\
  \cmark & \cmark & \cmark & \xmark & 45.51& 3.74\\
  \xmark & \cmark & \cmark & \cmark & 44.86 & 3.09 \\
     \cmark & \cmark & \cmark &\cmark &\textbf{46.00}&	4.23  \\
\bottomrule
\end{tabular}
\end{minipage}\hfill
\begin{minipage}[]{0.48\linewidth}
    \centering
    \includegraphics[width=\linewidth]{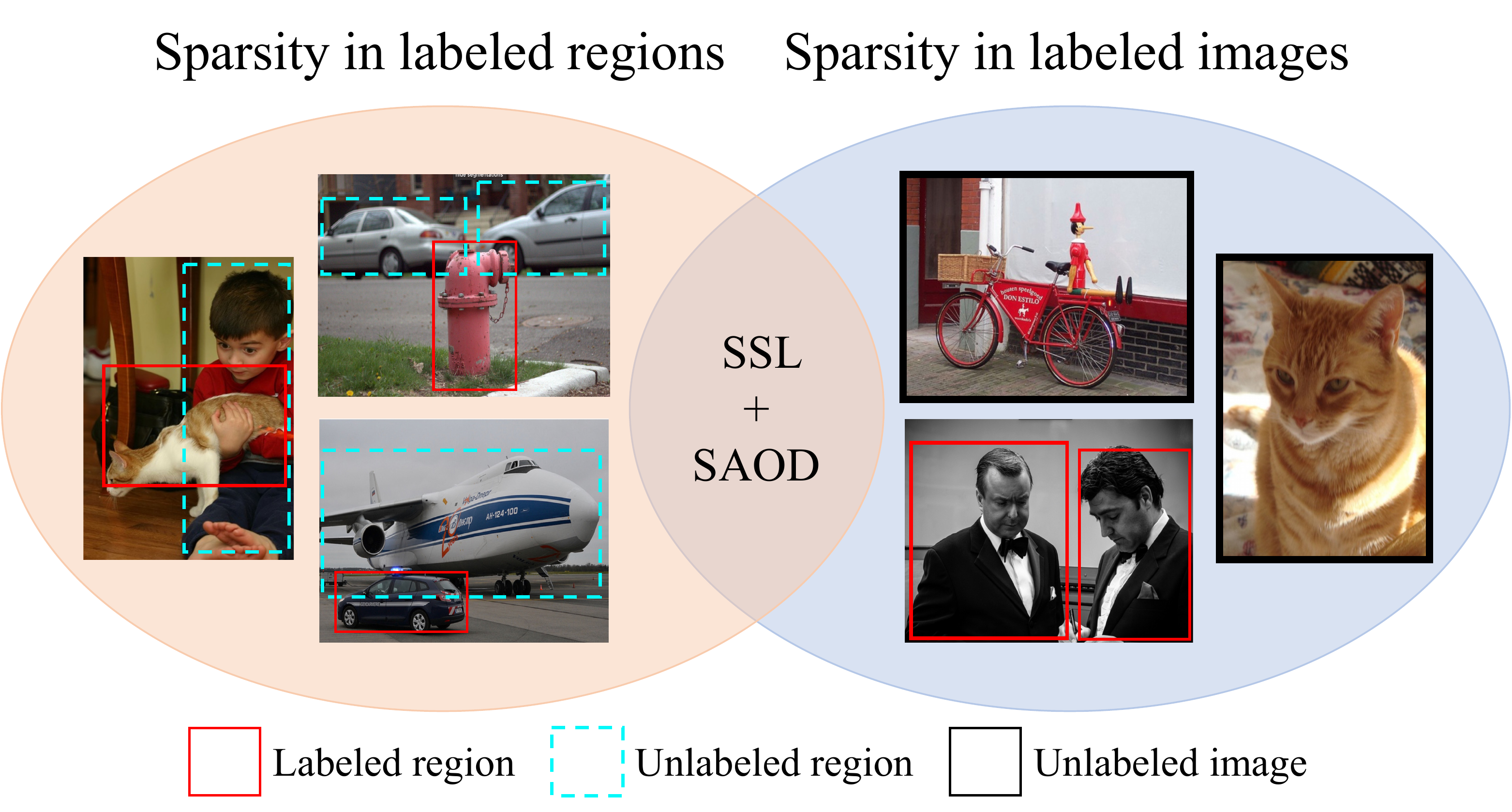}
    \captionof{figure}{Types of sparsity in labeled data; sparsity in labeled regions (left) and images (right). Our proposed SSL+SAOD is a realistic setup that presents challenges from both kinds of sparsity.}
    \label{fig:ssl_saod}
\end{minipage}\vspace{-1.2em}
\end{table*}

\subsection{Comparison with state-of-the-art}\label{sec:sota}
In this section, we compare our method with contemporary methods. We evaluate all the models in two different setups. We name the first setup \textit{Sparsely annotated} setup which evaluates models on SAOD. The second setup, contains labeled and unlabeled images and regions.\\

\noindent\textbf{\textit{Sparsely annotated} setup: }
We show results of this setup in Table~\ref{tab:saod_coco} and Table~\ref{tab:saod_voc}. Splits-2 and 3 ensure at least one annotation per image is retained. On the other hand, Split-1 doesn't ensure this and has significantly more unlabeled data than the other splits. Splits-4 and 5 have no unlabeled images at all.
In both the tables, the rows named ``oracle" refers to the models trained using all annotations.

From Table~\ref{tab:saod_coco}, we observe that our approach outperforms the other baselines and is closer to the oracle performance on the $30\%$ splits. Co-mining~\cite{Wang2021CominingSL}, performs competitively on all the splits at $30\%$. 

On Split-1, our method obtains a performance improvement of $1.87$, $3.08$ and	$7.75$ on $30\%$, $40\%$ and $50\%$ sparsity respectively over Co-mining~\cite{Wang2021CominingSL}. 

On Split-2, our method obtains an improvement of $3.04$,	$3.9$ and $10.5$ at $30\%$, $40\%$ and $50\%$ sparsity respectively over the previous state of the art Co-mining. 
On Split-3, an again we see consistent improvement of $2.8$, $4.61$ and $10.52$ at $30\%$, $40\%$ and $50\%$ sparsity respectively over Co-mining. 

In particular, on the hardest setting (70\%) in Split 1-3, we demonstrate that current SAOD methods struggle at higher sparsity in the labeled data. This is because the performance of current SAOD approaches rely on the quality of pseudo labels which degrades at higher sparsity. Our proposed PPM prevents penalizing the classifier irrespective of the quality of pseudo labels and utilizes self-supervised loss to benefit from mined regions. On average, all the methods achieve lower performance on Split-1 compared to the other splits due to the nature of its creation, \ie, class aware dataset level fashion of sparsity. 

From Table~\ref{tab:saod_voc}, we see a similar trend as above. All the methods perform competitively on the easier settings of both splits. However, the performance gap between the our approach and Co-mining~\cite{Wang2021CominingSL} increases at higher sparsity. 

On Split-4, we get an improvement of $2.56$, $3.12$ and $5.99$ (AP$_{50}$) percentage points on the \textit{easy}, \textit{hard} and \textit{extreme} settings respectively over Co-mining. On Split-5, an improvement of $2.42$, $2.58$ and $4.46$ points was observed. For details on more experiments refer to supplementary material. \\

\noindent\textbf{\textit{Single Instance Object Detection (SIOD)} setup}: \cite{li2022siod} proposed SIOD where only one instance per category is annotated in every image. While this setup has its benefits~\cite{li2022siod}, note that it is a special case of SAOD. For this setup, we obtain an AP of $32.89$ (compared to $31.9$ of~\cite{li2022siod}) which is an improvement of ${\sim}1$ mAP. \\

\begin{table*}[!h]
\begin{minipage}[b]{0.47\linewidth}

\renewcommand{\tabcolsep}{4pt}
\renewcommand{\arraystretch}{1.6}
\centering
\caption{Analysis of the threshold used for PPM.}
\label{tab:thresh}
\begin{tabular}{@{}lccccc@{}}
\toprule
 \textbf{Threshold} & 0.6 & 0.7 & 0.8& 0.9 & 0.95   \\
 \midrule
 \textbf{AP} & 45.96 & 45.96 & \textbf{46.00} & 45.77 & 45.30\\
\bottomrule
\end{tabular}
\end{minipage}
\begin{minipage}[b]{0.53\linewidth}
\renewcommand{\tabcolsep}{3pt}
\renewcommand{\arraystretch}{1.3}
\centering
\caption{Removing test time augmentations.}\label{tab:tta} %
\resizebox{\linewidth}{!}{
\begin{tabular}{lccccccccc}
\toprule
 \textbf{Method} & \multicolumn{3}{c}{\textbf{Split-1}}& \multicolumn{3}{c}{\textbf{Split-2}} & \multicolumn{3}{c}{\textbf{Split-3}} \\
 \cmidrule[\cmidrulewidth](l){2-4}
 \cmidrule[\cmidrulewidth](l){5-7}
 \cmidrule[\cmidrulewidth](l){8-10}
 &  30\% & 50\% & 70\% & 30\% &50\% & 70\% & 30\% &50\% & 70\% \\
 \midrule

 TTA  & 38.22 & 35.92 & 32.68& 39.76 & 36.94 & 35.33 & 39.56  & 37.15 & 35.48 \\ %
 w/o TTA & 38.32 & 35.91 & 32.67& 39.77 & 36.95 & 35.27 & 39.55  & 37.12 & 35.48 \\

\bottomrule
\end{tabular}
}
\end{minipage}
\vspace{-0.2em}
\end{table*}

\noindent\textbf{\textit{SSL+SAOD} setup: } We propose a semi-supervised learning benchmark for SAOD. This benchmark entails training models on a sparsely annotated labeled and an unlabeled set. As shown in Figure~\ref{fig:ssl_saod}, this setup introduces two kinds of sparsity in the data, namely, sparsity in labeled regions (left) and images (right). 
We believe this is a realistic setup and SAOD methods must be capable of tackling both these kinds of sparsity in the data.
For this setup, we use Split-5 (Section~\ref{subsec:split}) with increasing sparsity as the labeled set and VOC12 trainval as the unlabeled set. We use the COCO-style AP metric to report results on this setup. 

In Table~\ref{tab:tssl}, we compare against Co-mining~\cite{Wang2021CominingSL} and observe an improvement of $1.81$, $1.84$ and $3.55$ mAP on the 30\%, 40\% and 50\% sparsity levels respectively. We observe that the gap in performance increases with sparsity, consistent with our observation for Tables~\ref{tab:saod_coco} and~\ref{tab:saod_voc}. This can be attributed to the inability of methods like Co-mining to handle unlabeled images and high sparsity in the labeled data.
We will make this benchmark public and propose this method as a baseline. We encourage researchers to report results on this benchmark in the future.
\begin{figure*}[!t]
    \centering
\begin{minipage}[t]{\linewidth}
 \centering
\includegraphics[width=\linewidth]{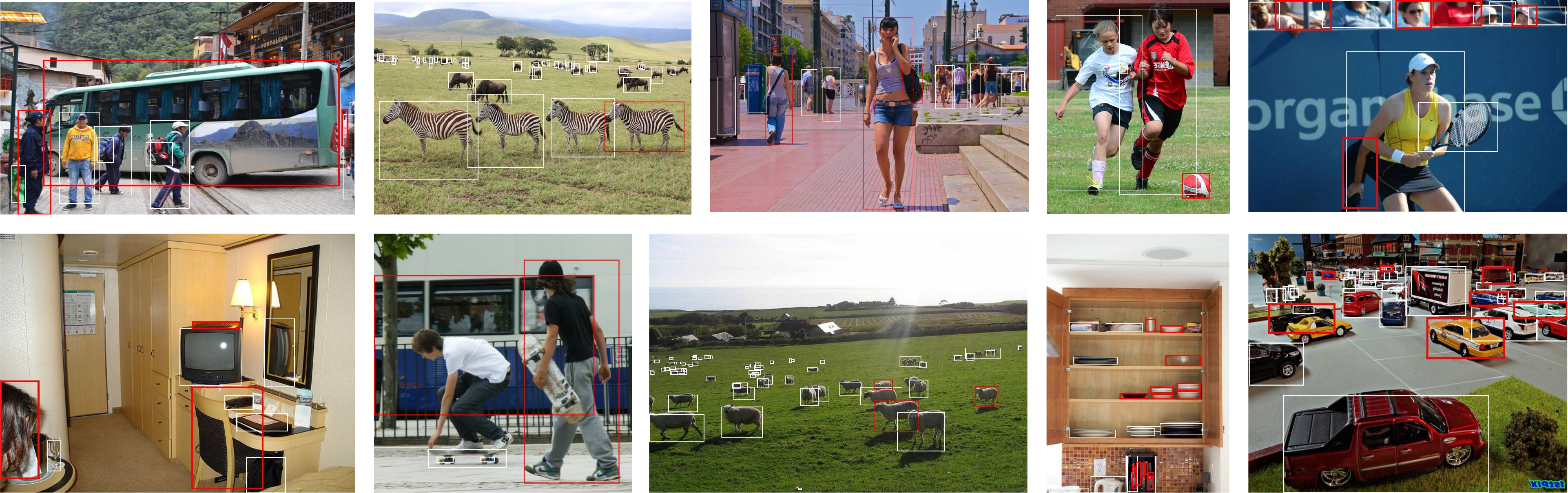}
\caption{Qualitative results showing the unlabeled regions identified by the PPM. The red boxes correspond to the available ground truth. A class agnostic NMS was performed on the regions and the result is shown in white. }\label{fig:ppm}
\end{minipage}
\begin{minipage}[t]{\linewidth}
\centering
\includegraphics[width=\linewidth]{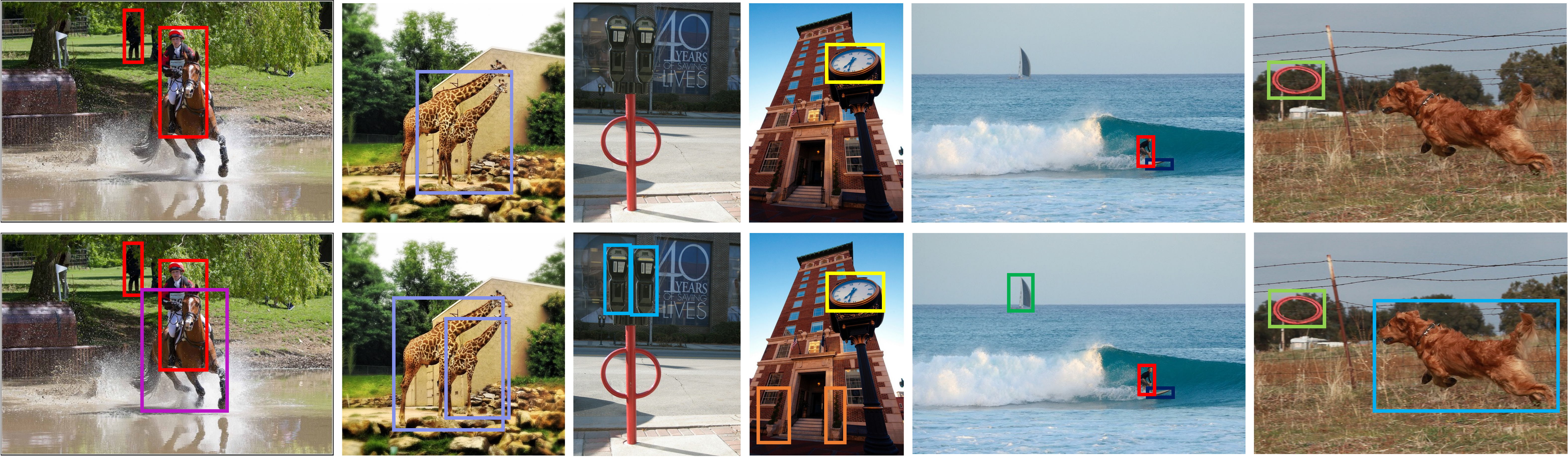}
    \caption{Qualitative results comparing the output of a model trained using available ground truths (top) to a model trained using our approach (bottom).  Predictions with a class confidence score greater than $0.9$ are shown. \textcolor{red}{Red}: Person, \textcolor{cyan}{Cyan}: Dog, \textcolor{purple}{Purple}: Horse, \textcolor{yellow}{Yellow}: Clock, \textcolor{YellowGreen}{Green}: Stop sign, \textcolor{Cerulean}{Blue}: Parking meter, \textcolor{violet}{Violet}: Giraffe, \textcolor{orange}{Orange}: Potted plant, Black: Surfboard, \textcolor{ForestGreen}{Dark green}: Boat}
    \label{fig:good_res}
\end{minipage}\vspace{-1.5em}

\end{figure*}

\subsection{Ablation Experiments}\label{sec:ablation}
In this section we conduct ablation experiments to understand the various components. For the ablation experiments, we use a ResNet-101 as the backbone network and train on Split-5 with $p = 50\%$. We evaluate on the VOC07 test set and report the COCO style AP metric.

From Table~\ref{tab:cr}, a baseline model trained on the ablation set attains $41.77$ (row 1). Pseudo-labeling $\left(\mathcal{L}^s_{\text{det},a}\right)$ improves the performance by $0.9$ (row 2). Co-mining~\cite{Wang2021CominingSL} relies extensively on pseudo-labels. This results in a drop in performance at higher sparsities due to noisy pseudo-labels.
Addition of the C-RPN improves our performance by $3.53$ points (row 3). C-RPN reduces the overhead of computing two sets of proposals and learns a better notion of objects due to the combined processing of features from the two views. 

The combination of C-RPN and PPM improves the performance by $3.67$ (row 4). We do not observe major improvements with the introduction of PPM because its task is to identify and segregate the unlabeled regions from the backgrounds. After the segregation, PPM does no further processing to improve performance. The power of PPM can be observed when trained in conjunction with  consistency regularization loss $\left(\mathcal{L}^\text{u}_{\text{SSL}}\right)$ which achieves the best performance of $46$ (row 7); an improvement of $4.23$ points over the baseline. 
We distinguish ourselves from pseudo-labeling approaches like Co-mining in one important aspect. While, Co-mining~\cite{Wang2021CominingSL} relies solely on pseudo-labels, we leverage additional components from self-supervised learning along with pseudo-labeling. In row 5, due to C-RPN, PPM and $\mathcal{L}^\text{u}_{\text{SSL}}$, we show an improvement of $3.74$ over the baseline. Our proposed components are orthogonal to pseudo-labeling as using them together results in an additional improvement of ${\sim}0.5$ on mAP. At higher sparsity in the labeled dataset, the proposed C-RPN, PPM and $\mathcal{L}^\text{u}_{\text{SSL}}$ are less affected than pseudo-labels resulting in the large improvements on these splits. Finally, we show the effect of C-RPN by generating a single set of proposals from the original image and using it for both the branches. We observe a drop in performance (row 6) highlighting the effectiveness of C-RPN.

\begin{table*}[t]
 \setlength{\cmidrulewidth}{0.01em}
\renewcommand{\tabcolsep}{3pt}
\renewcommand{\arraystretch}{1.1}
\centering
\caption{Comparison with a different backbone. Results are reported on the COCO validation set on Splits 1-3 using AP and on VOC 2007 test set on Splits-4,5 using AP$_{50}$.}\label{tab:backbone}
\resizebox{\linewidth}{!}{
\begin{tabular}{lccccccccccccccc}
\toprule
 \textbf{Method} &  \multicolumn{3}{c}{\textbf{Split-1}}&  \multicolumn{3}{c}{\textbf{Split-2}} & \multicolumn{3}{c}{\textbf{Split-3}} & \multicolumn{3}{c}{\textbf{Split-4}} & \multicolumn{3}{c}{\textbf{Split-5}}\\
 \cmidrule[\cmidrulewidth](l){2-4}
 \cmidrule[\cmidrulewidth](l){5-7}
 \cmidrule[\cmidrulewidth](l){8-10}
 \cmidrule[\cmidrulewidth](l){11-13}
 \cmidrule[\cmidrulewidth](l){14-16}

 &  30\% & 50\% & 70\% & 30\% & 50\% &70\% & 30\% & 50\% & 70\% & Easy & Hard & Extreme & 30\% & 40\% & 50\% \\
 \midrule
 Ours (C4) &  37.67 & 35.95 & 33.16 & 39.22 & 36.81 & 34.98 & 38.84 & 36.76 & 35.26 & 80.56 & 80.43 & 74.45 & 77.38 & 76.13 & 75.32 \\
 
 Ours (FPN) & 38.22 & 35.92 & 32.68 &	39.76 &	36.94 &	35.33 &	39.56 & 37.15 &	35.48 & 82.15 & 81.50 & 75.59& 76.84 & 75.88 & 74.35 \\
 
\bottomrule
\end{tabular}
}
\vspace{-1.2em}
\end{table*}

PPM mines potential positives which can be mistaken for negatives due to missing annotations. We rely on the objectness score of the RPN to identify these regions. In Table~\ref{tab:thresh}, we vary the threshold of PPM. For a low threshold, a few hard negatives might also be identified as pseudo-positive leading to a drop in performance. With a high threshold, a few potential positives might not be mined. We observe that a threshold of $0.8$ provides a good trade-off and is therefore used for all experiments unless stated otherwise. In all our experiments, PPM is performed after an initial warmup of $9000$ and $30000$ iterations on the PASCAL-VOC and COCO datasets respectively.

\subsubsection{Inference without augmentations}
Our method requires passing an input with augmentations during inference as well. It should be noted that this is not a test-time augmentation (TTA), a technique that typically involves passing images at a higher resolution. We perform inference by removing the augmentation and extracting the region proposals using C-RPN. We show the results in Table~\ref{tab:tta} on the three splits of COCO. We do not observe a significant improvement in performance due to the augmentation.
For the analysis on the effect of augmentations refer to supplementary material.
\subsubsection{Effect of backbone}
We analyze the effect of backbone on our approach and show results for the normal convolution based (C4) and FPN based (FPN) backbones in Table~\ref{tab:backbone}. The first row corresponds to C4 while the second row corresponds to FPN. While the gap in performance is lower in most cases, we observe a significant improvement using the FPN on the low sparsity settings of all the split. 

\subsection{RPN Recall experiments for object discovery}

PPM identifies foreground regions mistakenly assigned as background during training to avoid penalizing the network. To study the effect of PPM on novel~\cite{9093355} classes, we train a network using our approach on randomly chosen $6000$ images of the COCO dataset, containing annotations for only $20$ classes of PASCAL-VOC. We evaluate the recall$@0.5$ of the RPN over the remaining 60 classes. A model trained using the standard technique on this dataset achieves a recall$@0.5$ of $77.46\%$ and $29.06\%$ on the known classes (20 categories) and unknown classes (60 categories) respectively. Our proposed approach, with PPM, achieves a recall$@0.5$ of $78.47\%$ and $35.20\%$ respectively. This ability to localize objects not seen during training can be beneficial for object discovery methods like~\cite{Vo20rOSD,VoBCHLPP19,Rambhatla2021ThePO,MOST} which use RPN proposals to learn/discover new categories. 

\subsection{Qualitative Results}
In Figure~\ref{fig:ppm}, we show the pseudo positives mined by PPM. In each figure, the red boxes correspond to the ground truth annotations and the white boxes correspond to the post NMS pseudo positive boxes mined by PPM. We observe that the PPM correctly mines proposals which correspond to missing object annotations. Without PPM, these regions will be used as negatives to the classifier resulting in a reduced class confidence score leading to a drop in mAP.
We show the detection results of a model trained on the 50\% Split-1 of our approach in Figure~\ref{fig:good_res}. The images on the top corresponds to the model trained using sparse annotations and the bottom image shows the output of our approach.

%% file: conclusion.tex
\section{Conclusion}
\label{sec:conclusion}
We present SparseDet, a novel end-to-end system for Sparsely Annotated Object Detection (SAOD). We propose a simple yet effective technique for identifying the unlabeled regions using pseudo-positive mining and apply self-supervised loss on them. Through qualitative results we highlight the ability of PPM to mine pseudo-positives.
We standardize the evaluation setup and show the effectiveness of our approach with an exhaustive set of experiments on multiple splits of SAOD. While we outperform existing state-of-the-art on all metrics and splits, we observe the gap in performance increases with sparsity demonstrating the short coming of methods that rely solely on pseudo-labeling. We propose a new benchmark, that evaluates the semi-supervised learning capabilities of SAOD approaches. We will release the data for the new benchmark along with all the SAOD splits and encourage researchers to evaluate future SAOD methods on these.

%% file: acknowledgement.tex
\vspace{-0.1in}
\begin{small}
\paragraph{Acknowledgements.} This project was partially funded by DARPA SAIL-ON (W911NF2020009), DARPA SemaFor (HR001119S0085), and IARPA SMART (2021-21040700003) programs. Rama Chellappa was supported by an ONR MURI Grant N00014-20-1-2787. We would like to thank our colleagues from the PI lab for their feedback on this work. The views and conclusions contained herein are those of the authors and do not represent views, policies, and/or endorsements of funding agencies.
\end{small}

%% file: iccv_supp.tex
\setcounter{section}{0}

\twocolumn[{%
 \centering
 \Large Supplementary material for SparseDet: Improving Sparsely Annotated Object Detection with Pseudo-positive Mining\\[1.5em]
}]
\section{Recall analysis}
To study the effect of missing annotations on the RPN, we compute the recall of the RPN trained using a sparsely annotated dataset. For a model trained using all the annotations, the recall of the RPN is $0.83$.
As we drop annotations progressively from 30\% to 70\% the recall drops to $0.79$.  This shows that there is no significant drop in recall due to missing annotations.

\begin{table}[!ht]
 \setlength{\cmidrulewidth}{0.01em}
\renewcommand{\tabcolsep}{4pt}
\renewcommand{\arraystretch}{1}
\centering
\caption{Ablation for $\tau_m$}
\label{tab:tm}
\begin{tabular}{@{}lccc@{}}
\toprule
 \textbf{Threshold} & 0.8& 0.9 & 0.95   \\
 \midrule
 \textbf{AP}  & 45.04 &\textbf{46.00}  & 45.38\\
\bottomrule
\end{tabular}
\end{table}
\section{Additional Experiments}

\subsection{Experiments on publicly released splits}
Authors of Co-mining~\cite{Wang2021CominingSL} create a split similar to Split-4 but using COCO 2017 trainval set to perform ablation experiments. To the best of our knowledge, this is the only split released publicly for SAOD. We compare the performance of our approach with Co-Mining, which uses a RetineNet architecture, on this split and report results in Table \ref{tab:cosplit}. We observe a similar trend as reported in the main paper and outperform Co-mining by $1.38$, $2.22$ and $0.35$ percentage points on the \textit{Easy}, \textit{Hard} and \textit{Extreme} subsets respectively.

\subsection{Effect of $\tau_m$ }
We ablate over different values of $\tau_m$ and report the results in Table~\ref{tab:tm} . With a low threshold, the ground truth will be contaminated with noisy predictions and a high threshold will leave out a few positive missing annotations. We observed that a value of 0.9 is a good tradeoff between quality and recall. We use a threshold of 0.9 for all the datasets unless otherwise stated.
\subsection{Warm-up}
As we rely on the RPN objectness score, we employ a warmup strategy which switches on the pseudo-positive mining (PPM) during training. In Table \ref{tab:warmup} we start the PPM at different iterations and report results. This experiment is performed on the COCO-mini ablation dataset. Our experiments show that for models which are trained for fewer iterations, starting at $9000$ iterations worked the best. For longer experiments we allow a warmup of $30000$ iterations.

\subsection{Additional ablations on C-RPN}
We perform some additional analysis regarding C-RPN by using two sets of proposals, one from original and the other from augmented version of the image and combining them. This approach consists of $2\times$ more proposals than our approach and achieves an mAP of 45.25 (vs.\ \textbf{46.00} using ours). To make the comparison fairer, we generate half the number of proposals from each image and combine them (resulting in the same number of proposals as our approach) and this model achieves 45.46 (vs.\ \textbf{46.00} using ours) demonstrating the effectiveness of C-RPN.

\subsection{Augmentation}
The proposed approach process an input image and its augmented counterpart. For augmentation, a cascade of random contrast, brightness, saturation, lighting and bounding box erase are used. In Table \ref{tab:aug}, we analyze the effect of various augmentations on the performance of the model. This experiment was conducted on the same ablation set as the main paper. We observe that applying random saturation or random lighting alone are not as effective compared to other augmentations. Applying bounding box erase alone provides the most improvement. Finally, we achieve the best performance when we use all the augmentations. We randomly sample contrast, brightness, and saturation from $\left[0.5, 1.5\right]$. For lighting, a random scale of $1.2$ was used. For bounding box erase, we randomly erase an area of of $\left[0.4, 0.7\right]$ at an aspect ratio of $\left[0.3, 3.3\right]$.

\begin{table*}[t]
\begin{minipage}[t]{0.43\linewidth}
\begin{minipage}[t]{\linewidth}
\setlength{\cmidrulewidth}{0.01em}
\renewcommand{\tabcolsep}{10pt}
\renewcommand{\arraystretch}{1.1}
\caption{Results on the splits released by authors of \cite{Wang2021CominingSL}. All methods use Res-50 FPN architecture.}\label{tab:cosplit}
\resizebox{\linewidth}{!}{
\begin{tabular}{@{}lccc@{}}
\toprule
 \textbf{Method}& \multicolumn{3}{c}{\textbf{Performance}} \\
 \cmidrule{2-4}
 &  \textit{Easy} & \textit{Hard} & \textit{Extreme}\\
 \midrule
 Co-mining\cite{Wang2021CominingSL} & 35.40 & 31.80 & 23.00 \\
 Ours & \textbf{36.78} & \textbf{34.02} & \textbf{23.35}\\
\bottomrule
\end{tabular}
}
\end{minipage}\vspace{0.5em}

\begin{minipage}[t]{\linewidth}
\renewcommand{\tabcolsep}{8pt}
\renewcommand{\arraystretch}{1.1}
\caption{Comparison with ~\cite{niitani2019sampling} on Split-2}
\label{tab:s2}
\resizebox{\linewidth}{!}{
\begin{tabular}{@{}lccc@{}}
\toprule
 \textbf{Method} &\multicolumn{3}{c}{\textbf{Performance}} \\
 \cmidrule{2-4}
 &  \textit{30\%} & \textit{50\%} & \textit{70\%}\\
 \midrule
 Pseudo Label~\cite{niitani2019sampling}   & 35.00 & 32.79 & \textbf{29.03} \\
  Soft Sampling~\cite{wu2018soft}   & 33.98 & 31.39 & 27.30 \\
 Ours  & \textbf{35.94} & \textbf{33.13} & 28.88 \\
\bottomrule
\end{tabular}
}
\end{minipage}
\end{minipage}\hfill
\begin{minipage}[t]{0.52\linewidth}
\setlength{\cmidrulewidth}{0.01em}
\renewcommand{\tabcolsep}{4pt}
\renewcommand{\arraystretch}{1.1}
\caption{Analysis of different augmentations used in this work.}\label{tab:aug}
\resizebox{\linewidth}{!}{
\begin{tabular}{@{}cccccc@{}}
\toprule
 \textbf{Contrast}& \textbf{Brightness} & \textbf{Saturation} & \textbf{Lighting}&\textbf{Bbox-Erase} & \textbf{Performance}  \\
 \midrule
 \xmark & \xmark & \xmark & \xmark & \xmark & 42.76\\
 \cmark & \xmark & \xmark & \xmark & \xmark & 43.99  \\
 \xmark & \cmark & \xmark &\xmark & \xmark & 43.81 \\
  \xmark & \xmark & \cmark & \xmark & \xmark & 42.70 \\
 \xmark & \xmark & \xmark &\cmark & \xmark & 42.95\\
  \xmark & \xmark & \xmark & \xmark & \cmark & 45.11  \\
  \midrule
 \cmark & \cmark & \xmark &\xmark & \xmark & 44.03 \\
  \cmark & \cmark & \cmark & \xmark & \xmark & 43.90  \\
  \cmark & \cmark & \cmark & \cmark & \xmark & 43.91  \\
 \cmark & \cmark & \cmark &\cmark & \cmark & 46.00 \\
\bottomrule
\end{tabular}
}

\setlength{\cmidrulewidth}{0.01em}
\renewcommand{\tabcolsep}{12pt}
\renewcommand{\arraystretch}{1.1}
\centering
\footnotesize
\caption{Analysis on warm-up iteration for performing PPM}
\label{tab:warmup}\vspace{0.3em}
 \resizebox{\linewidth}{!}{
\begin{tabular}{@{}lcccc@{}}
\toprule
 \textbf{Iteration}&  0 & 4000 & 9000 & 12000  \\
 \midrule
 AP & 45.51 & 45.71 & \textbf{46.00} & 45.88\\
\bottomrule
\end{tabular}
}
\end{minipage}
\end{table*}

\section{Additional details of splits and experiments}
In Table \ref{tab:s2}, we compare our approach with Niitani \etal~\cite{niitani2019sampling} and Wu \etal~\cite{wu2018soft} using the same backbone for fair comparison. 
We outperform the best model~\cite{niitani2019sampling} by $0.94$, $0.34$ points on the $30\%$ and $50\%$ splits respectively.

\begin{figure*}
 \centering
\includegraphics[width=0.9\linewidth]{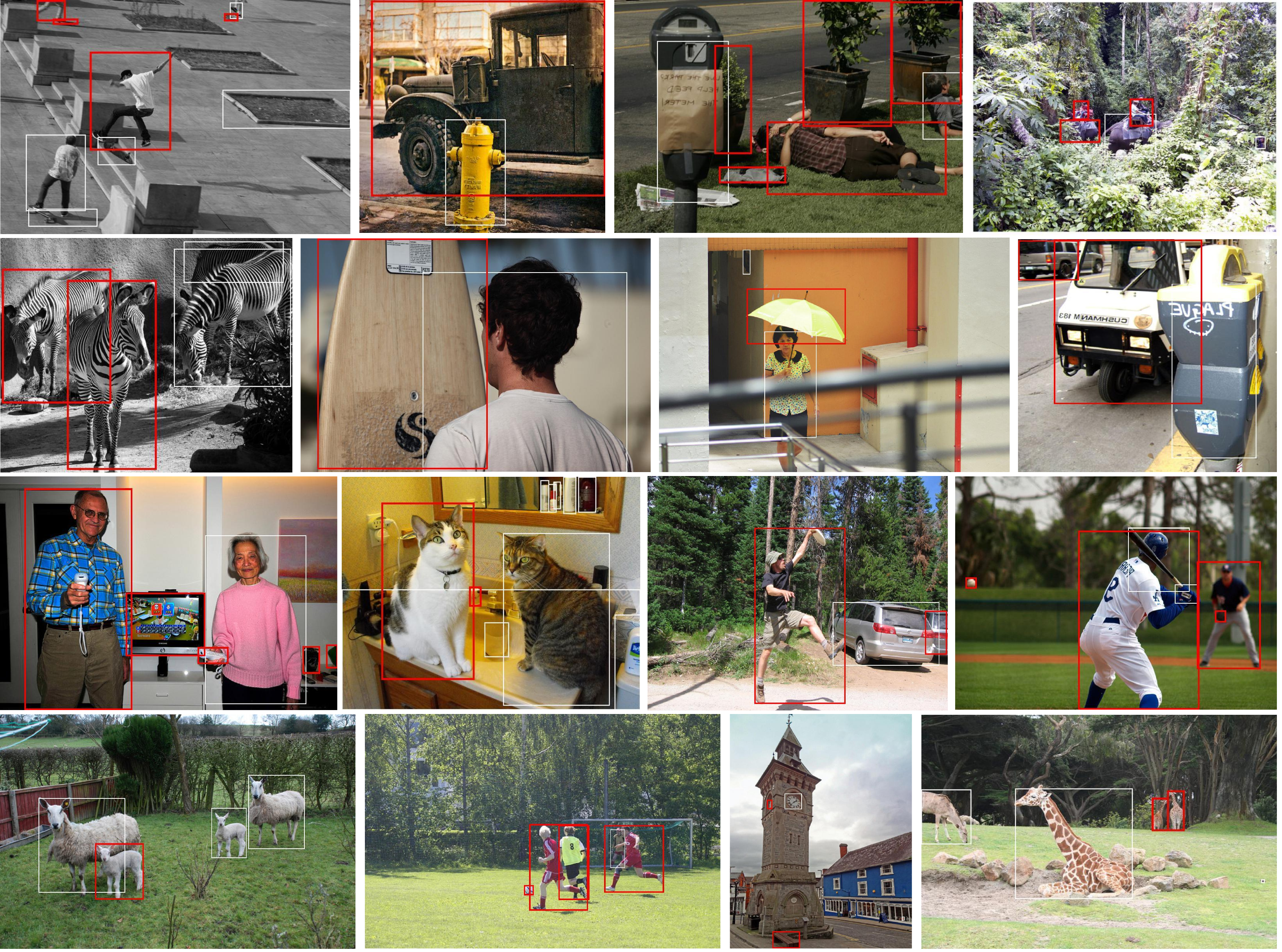}
\caption{Figure showing regions (white) identified by the PPM step and available ground truth (red) during training.}\label{fig:sup_ppm_pos}
\end{figure*}
\begin{figure*}
    \centering
    \includegraphics[width=0.9\linewidth]{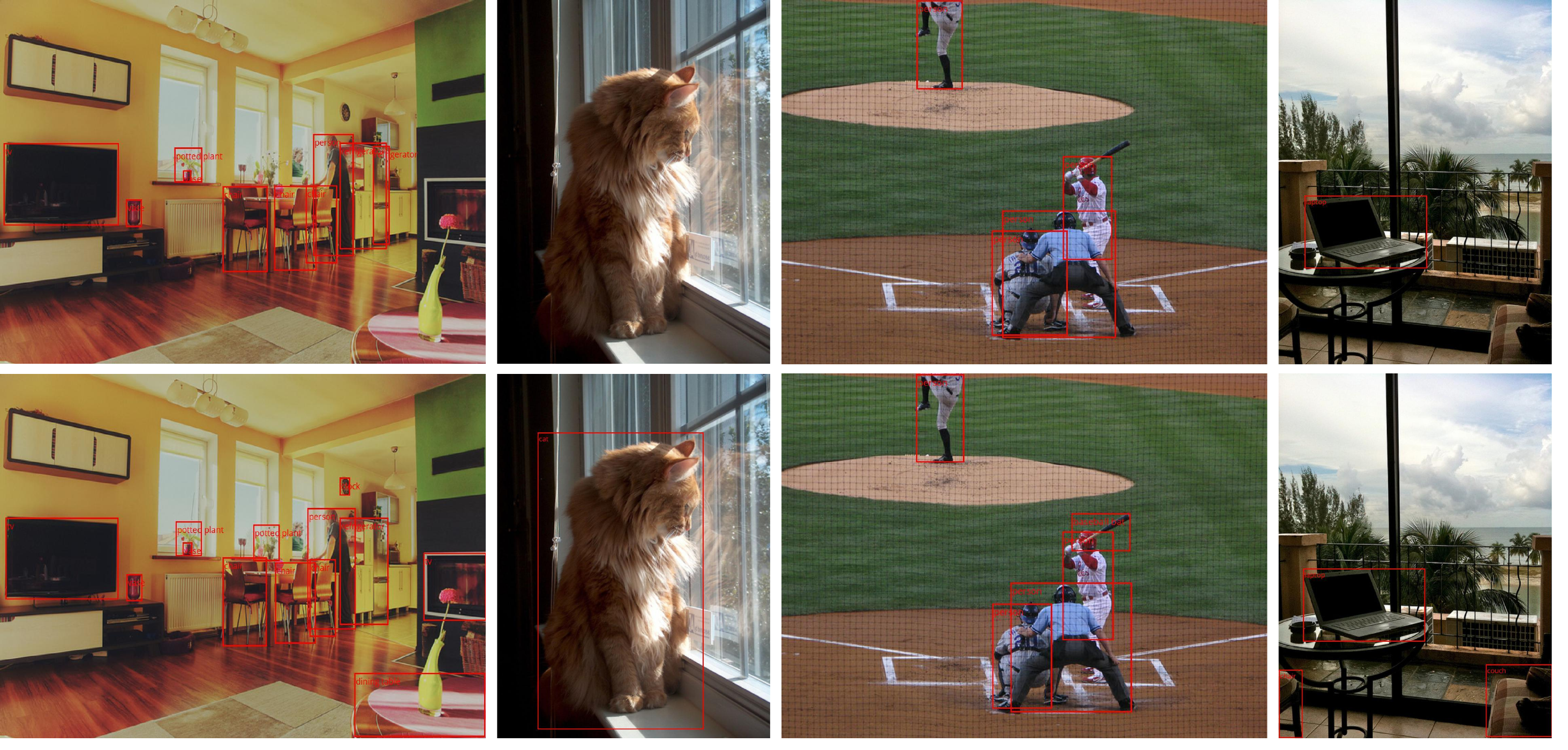}
    \caption{Qualitative results comparing the output of a model trained using available ground truths (top) to a model trained using our approach (bottom). Predictions with a class confidence score greater than 0.9 are shown.}
    \label{fig:det_pos}
\centering
\includegraphics[width=0.9\linewidth]{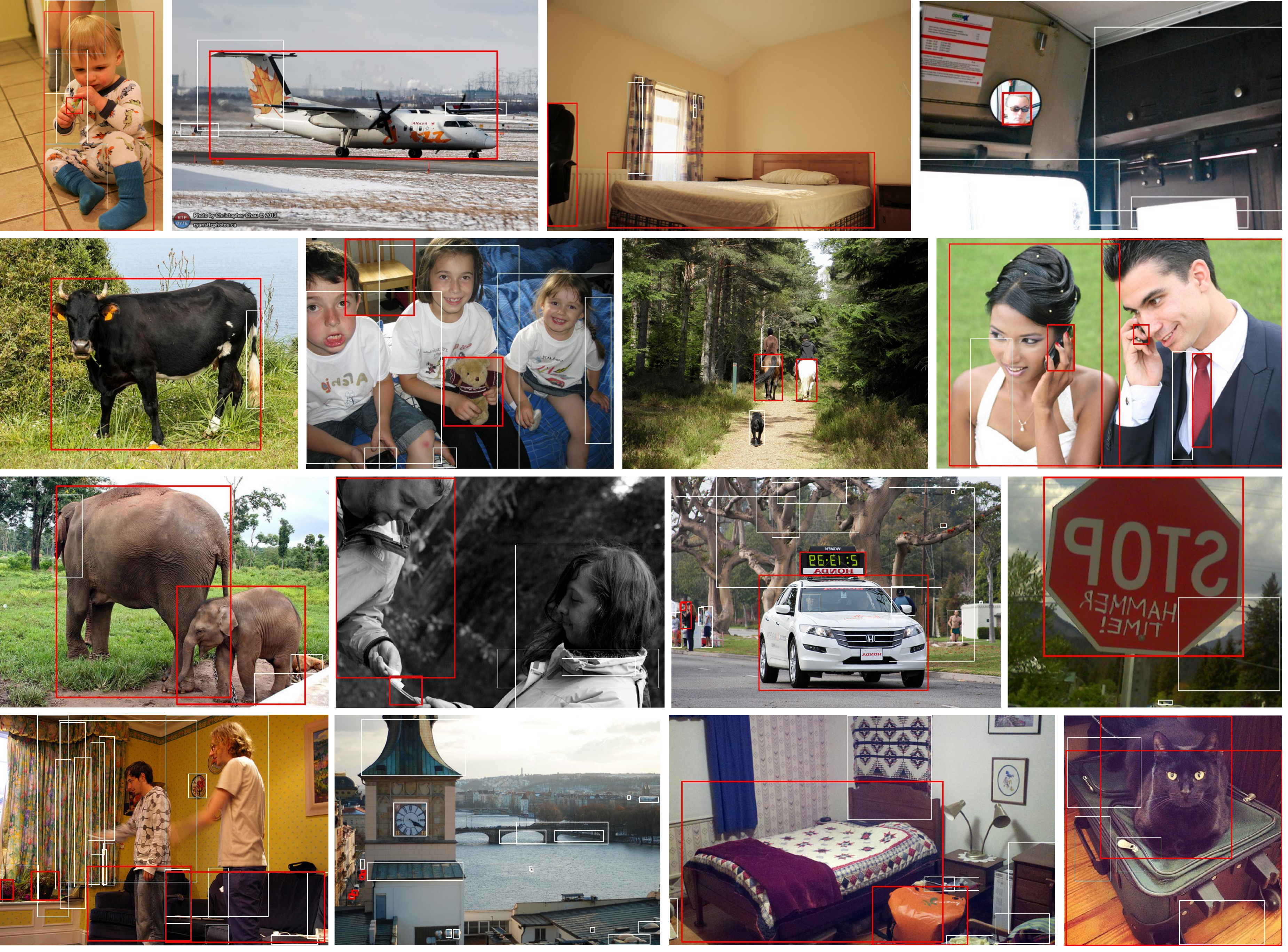}
    \caption{Failure cases of PPM.}
    \label{fig:sup_ppm_neg}\vspace{0.5em}
    \end{figure*}
 
\section{Results}
\begin{figure*}[t]
    \centering
    \includegraphics[width=0.9\linewidth]{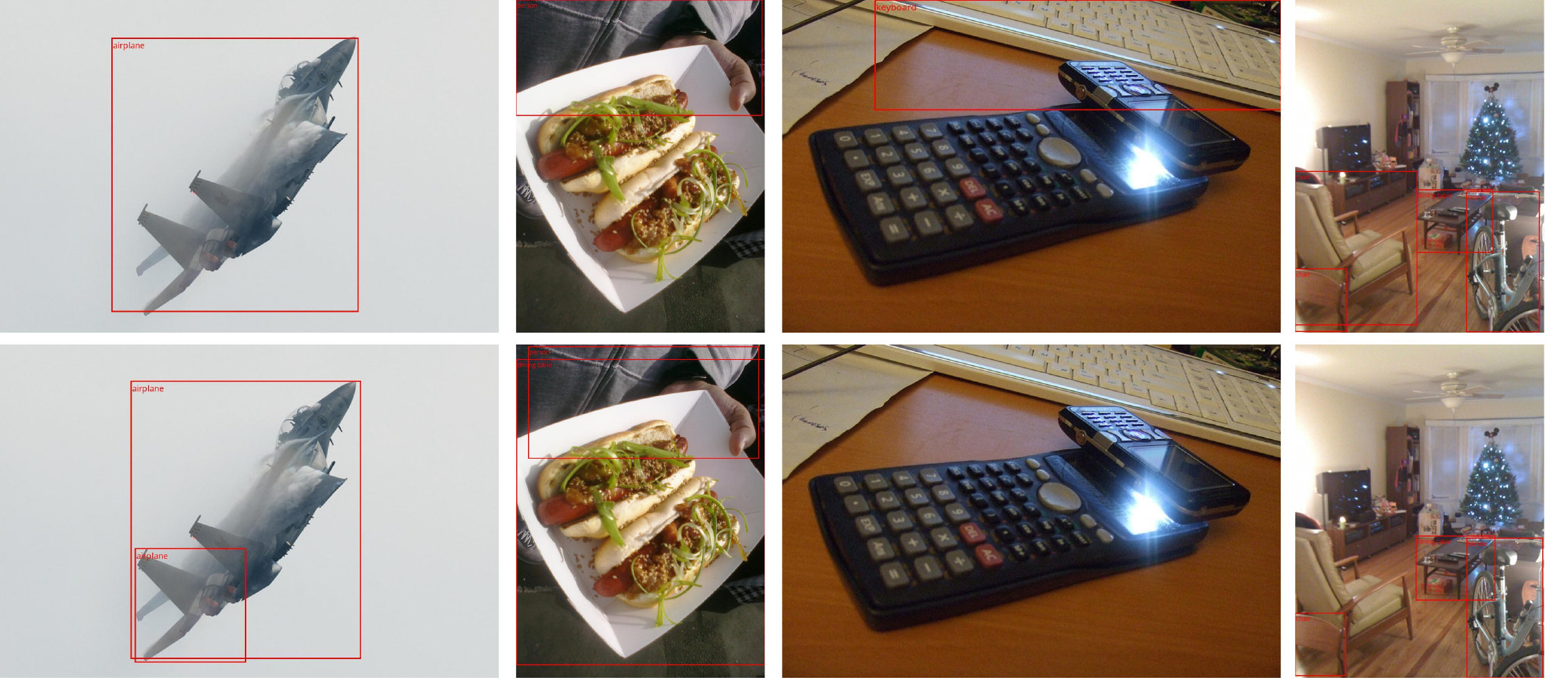}
    \caption{Failure cases comparing the output of a model trained using available ground truths (top) to a model trained using our approach (bottom). Predictions with a class confidence score greater than 0.9 are shown.}
    \label{fig:det_neg}
\end{figure*}

In this section, we show some more results of the regions identified by the PPM in Figure \ref{fig:sup_ppm_pos} and a few failure cases in Figure \ref{fig:sup_ppm_neg}. In Figures \ref{fig:sup_ppm_pos}, \ref{fig:sup_ppm_neg} the red boxes correspond to the available annotations and the regions in white are the ones identified by PPM. Finally, we show detection results and failures of our approach in Figures \ref{fig:det_pos} and \ref{fig:det_neg} respectively. 

